\documentclass{article}

\usepackage{setspace}
\usepackage{PRIMEarxiv}
\usepackage{hyperref}       % hyperlinks
\usepackage{amsfonts}       % blackboard math symbols
\usepackage{nicefrac}       % compact symbols for 1/2, etc.
\usepackage{fancyhdr}       % header
\usepackage{graphicx}       % graphics

\usepackage{adjustbox}
\usepackage{amsmath}
\usepackage{bbm}
\usepackage{algorithm}
\usepackage{algorithmic}
\usepackage[caption=false, font=footnotesize]{subfig}
\usepackage{caption}
\usepackage{multirow}
\usepackage[normalem]{ulem}
\useunder{\uline}{\ul}{}
\usepackage{xurl}
\urldef{\technical}\url{http://test-10056879.file.myqcloud.com/10056879/test/20180524_78431960010324/KPI%E5%BC%82%E5%B8%B8%E6%A3%80%E6%B5%8B%E5%86%B3%E8%B5%9B%E6%95%B0%E6%8D%AE%E9%9B%86.zip}

\pagestyle{fancy}
\thispagestyle{empty}
\rhead{ \textit{ }} 

\fancyhead[LO]{Multi-Task Self-Supervised Time-Series Representation Learning}
\title{Multi-Task Self-Supervised \\ Time-Series Representation Learning}

\author{
  Heejeong Choi, Pilsung Kang* \\
  School of Industrial \& Management Engineering \\
  Korea University \\
  Seoul \\
  \texttt{\{heejeong\_choi, pilsung\_kang\}@korea.ac.kr} 
}

\begin{document}
\setstretch{1.25}
\maketitle

\begin{abstract}
Time-series representation learning can extract representations from data with temporal dynamics and sparse labels. When labeled data are sparse but unlabeled data are abundant, contrastive learning, i.e., a framework to learn a latent space where similar samples are close to each other while dissimilar ones are far from each other, has shown outstanding performance. This strategy can encourage varied consistency of time-series representations depending on the positive pair selection and contrastive loss. We propose a new time-series representation learning method by combining the advantages of self-supervised tasks related to contextual, temporal, and transformation consistency. It allows the network to learn general representations for various downstream tasks and domains. Specifically, we first adopt data preprocessing to generate positive and negative pairs for each self-supervised task. The model then performs contextual, temporal, and transformation contrastive learning and is optimized jointly using their contrastive losses. We further investigate an uncertainty weighting approach to enable effective multi-task learning by considering the contribution of each consistency. We evaluate the proposed framework on three downstream tasks: time-series classification, forecasting, and anomaly detection. Experimental results show that our method not only outperforms the benchmark models on these downstream tasks, but also shows efficiency in cross-domain transfer learning.
\end{abstract}

% keywords can be removed

\section{Introduction}
Time-series analysis aims to extract meaningful knowledge from raw time-series data \cite{esling2012time}. As large-scale time-series data can be easily accessed in various domains from the Internet of Things (IoT), time-series analysis plays an essential role in a wide range of applications \cite{silva2018speeding}. In particular, decision support based on time-series analysis is becoming necessary for digital transformation in many companies and organizations \cite{soofastaei2020digital}. However, many real-world time-series data need to be labeled and require domain experts for complex annotations \cite{ching2018opportunities}. Therefore, it is very challenging to analyze a limited amount of labeled time-series data using deep learning, which relies heavily on a large number of training data to avoid overfitting \cite{wen2020time}.

Self-supervised learning has attracted considerable attention as it avoids the extensive cost of annotating large-scale datasets. It adopts self-defined pseudo-labels as supervised learning and uses the learned representations for downstream tasks with limited labeled data \cite{jaiswal2020survey}. Self-supervised representation learning has been extensively studied in various domains such as computer vision and natural language processing \cite{doersch2015unsupervised, kim2021self, komodakis2018unsupervised, kong2019mutual, noroozi2016unsupervised, shi2021simple}. However, only a few researches have been proposed for time-series analysis \cite{franceschi2019unsupervised, tonekaboni2021unsupervised, zerveas2021transformer, eldele2021time, yue2022ts2vec}.

The goal of time-series representation learning is to learn representations by capturing inherent information such as temporal dynamics and variable correlations in time-series data. It has been studied in two main categories: 1) pretext task-based approach and 2) contrastive learning-based approach. The former proposed various pretext tasks in which the networks learn representations on data with pseudo-labels automatically generated based on time-series attributes. TST \cite{zerveas2021transformer} trained the model on the pretext task where the network predicts the masked values in sufficiently long masked sequences. However, it has a limitation that learned representations from these approaches are difficult to generalize for other downstream tasks that are less related to the pretext task. The latter has become a dominant component in time-series representation learning to resolve the discrepancy between the pretext task and the downstream task. The contrastive learning-based approach explores consistency in time-series data by maximizing the similarity between the positive pairs and maximizing the dissimilarity between negative pairs. There are some variations with regard to the way of constructing contrastive pairs and defining contrastive loss. T-Loss \cite{franceschi2019unsupervised} mainly pursued the sub-series consistency that encourages representations of the input time segment and its sampled sub-series to be close to each other. TNC \cite{tonekaboni2021unsupervised} enforced temporal consistency of two adjacent time segments sampled from a temporal neighborhood using stationary properties. TS-TCC \cite{eldele2021time} explored transformation consistency by minimizing the distance between representations of two augmented views from a single time segment. TS2Vec \cite{yue2022ts2vec} proposed a contextual consistency, which treats the same timestamps in two overlapping time segments as positive pairs.

%Figure 1
\begin{figure*}[!t]
    \centering
    \includegraphics[width=0.9\textwidth]{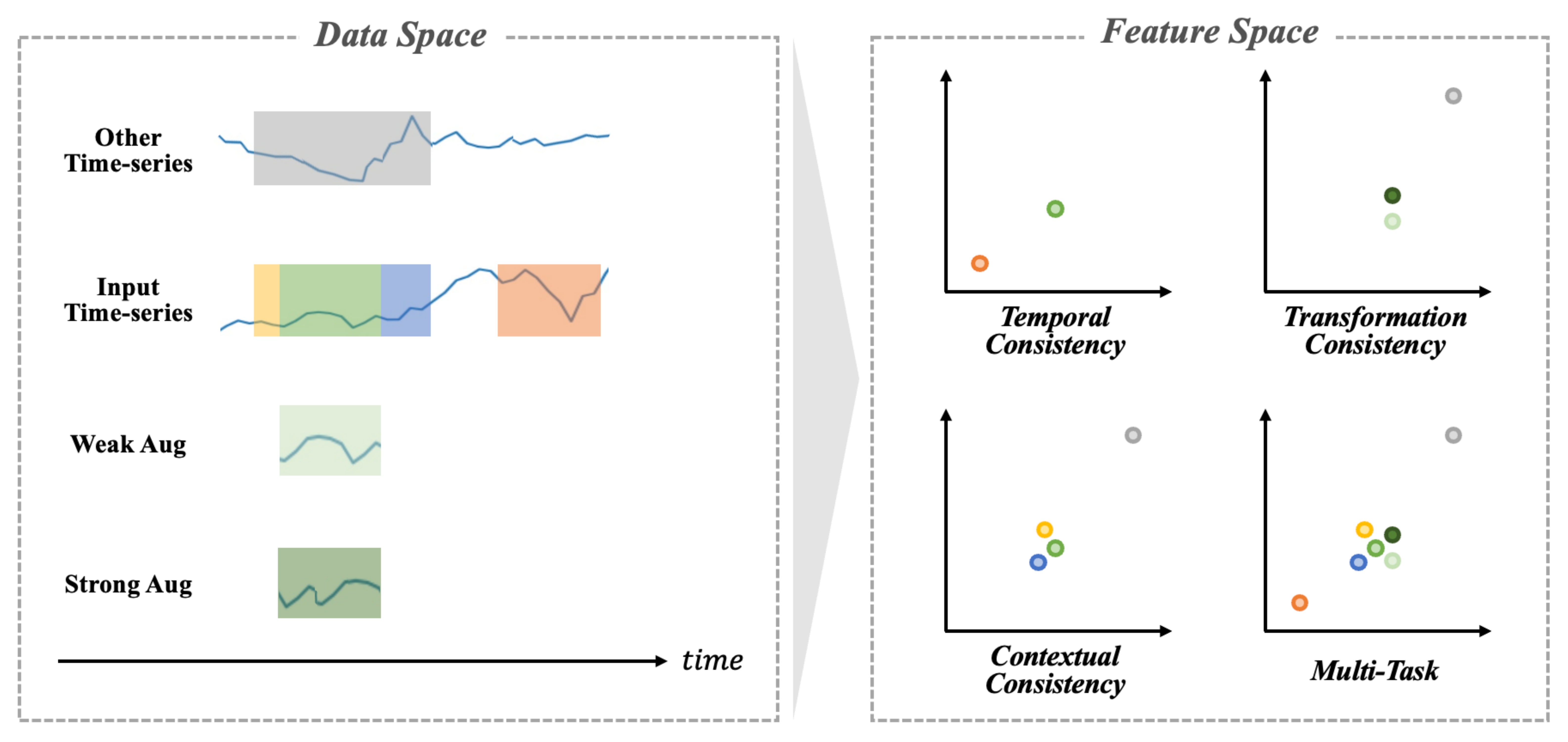}
    \caption{Feature space learned as per each consistency}
    \label{fig1}
\end{figure*}

Figure \ref{fig1} shows that the setting of contrastive learning allows the model to learn various relationships between time-series data. The left part shows the original time-series data while the right part illustrates the feature space where the model maps timestamps in each colored region. TNC with temporal consistency can learn representations whereby stationary and non-stationary regions in input time series are distinguished in the feature space. TS-TCC with transformation consistency can learn representations robust to perturbation and distinct from other instances. Furthermore, TS2Vec with contextual consistency can learn representations of timestamps around overlapping regions to near and ones of other instances to far away from them. Although each consistency derives distinct but useful information from time-series data, existing contrastive learning-based approaches lack the richness of the learned representations because they only focus on a single type of consistency. For example, the learned representations by temporal consistency could be weak in mapping perturbed inputs into feature space and reflecting contextual information. Therefore, a representation learning method is needed to comprehensively extract the features of time-series data from various perspectives by integrating various consistencies.

In this paper, we propose a multi-task self-supervised time-series representation learning framework to combine the advantages of contextual, temporal, and transformation consistency. First, contextual consistency \cite{yue2022ts2vec} hierarchically discriminates positive and negative samples at the instance and timestamp levels to capture contextual information at multiple resolutions. Second, temporal consistency \cite{tonekaboni2021unsupervised} encourages the local smoothness of representations by distinguishing time segments within the neighborhood from non-neighborhood ones. Third, transformation consistency \cite{eldele2021time} enforces the model to learn transformation-invariant representations by choosing time segments through two transformations as positive samples. Finally, we simultaneously learn these three contrastive learning approaches to model various consistencies together in time-series data. To jointly learn various self-supervised tasks effectively, we investigate an uncertainty weighting approach \cite{kendall2018multi} that can weigh multiple contrastive loss functions by considering the homoscedastic uncertainty of each task. We conducted downstream evaluations on time-series classification, forecasting, and anomaly detection. Experimental results demonstrated that our model could learn general representations for various downstream tasks. Furthermore, we performed transfer learning experiments and validated that the learned representations of our method are general for various domains.

The remainder of this paper is organized as follows. In Section 2, we briefly review related works. Section 3 describes our proposed model. Section 4 explains the experimental settings, followed by the experimental results. Finally, we summarize our study and list some future work directions in Section 5.

\section{Related Work}
\subsection{Self-Supervised Learning}
Self-supervised learning can learn features from large-scale unlabeled data without human annotation. Self-supervised learning was originally proposed in computer vision and has been studied from the perspective of the pretext task and contrastive learning. The mainstream of pretext task-related studies is as follows. Exemplar \cite{alexey2015discriminative} generated pseudo-labels where each class was constituted using a randomly sampled image patch and its transformed versions. The network was trained to discriminate representations between a set of pseudo-labels. Context prediction \cite{doersch2015unsupervised} learned visual features by predicting the relative positions among randomly extracted two patches from one image. Jigsaw puzzles \cite{noroozi2016unsupervised} aimed to learn spatial arrangement by solving an image patch jigsaw puzzle as the pretext task. Rotation \cite{komodakis2018unsupervised} transformed the images by the four image rotations and trained the network on the classification of recognizing the correct rotation. Colorization \cite{zhang2016colorful} transformed RGB images to grayscale ones and recovered the actual ground truth color from grayscale images. Although these approaches have successfully improved self-supervised visual learning, they can limit the generalization ability of the learned image features because they are more suitable for solving pretext tasks than downstream tasks.

Various contrastive learning frameworks have been proposed for self-supervised visual learning. MoCo \cite{he2020momentum} sampled queries and keys from images and built a dynamic dictionary as a queue of these data samples. Momentum encoder was trained to learn visual representations on discrimination of matching an encoded query to a dictionary of encoded keys using a contrastive loss. SimCLR \cite{chen2020simple} simplified contrastive learning framework without specialized architectures or memory banks by using a large batch of negative pairs. It further composed multiple data augmentations for effective positive pairs and introduced a projection head to improve the quality of the learned representations. SimSiam \cite{grill2020bootstrap} achieved good performance without negative pairs and momentum encoders. This model maximized the similarity between two augmentations of a single image using a Siamese network and stop-gradient operation. Furthermore, many self-supervised learning methods have improved the performance of computer vision tasks using limited labeled data \cite{larsson2017colorization, zhang2017split, pathak2016context, xie2016unsupervised, bojanowski2017unsupervised, caron2018deep, noroozi2017representation, noroozi2018boosting, chen2019self}, but it is difficult to apply these methods to time-series data owing to different characteristics of time-series data from images.

\subsection{Self-Supervised Time-Series Representation Learning}
Two mainstream approaches of self-supervised time-series representation learning are (1) employing an effective pretext task or (2) adopting a contrastive learning strategy. The TST \cite{zerveas2021transformer}, a pretext task-based method, is a Transformer-based framework that extracts dense representations by predicting the masked values of input time-series data. However, the pretext tasks can limit the generality of learned time-series representations. To address this issue, contrastive learning has been highlighted in time-series representation learning. T-Loss \cite{franceschi2019unsupervised} proposed an unsupervised triplet loss employing time-based negative sampling. It considered a sub-segment belonging to the input time segment as a positive sample to explore sub-series consistency, while introducing negative samples randomly chosen from other instances. TNC \cite{tonekaboni2021unsupervised} defined temporal neighborhood as a stationary region and enforced temporal consistency by distinguishing signals within it from non-neighboring signals. This method selected positive and negative samples from the neighborhood and non-neighborhood, respectively. TS-TCC \cite{eldele2021time} built two contrasting modules upon weak and strong augmentations to retain transformation consistency. It considered two augmented views as positive samples while treating augmented views from other instances in the batch as negative ones. In the first contrasting module, a cross-view prediction was performed to learn robust representations. The second module encouraged the model to learn discriminative representations by maximizing the similarity among representations of two augmented views. TS2Vec \cite{yue2022ts2vec} captured the contextual consistency by performing two contrastive learnings over two augmented time segments with different contexts. In both contrastive learnings, representations at the same timestamp in two augmented segments were considered positive pairs. In contrast, negative samples were represented differently in both augmented segments for the first contrastive learning, while representations at the same timestamp in other instances were treated as negative samples. Although these methods can learn time-series representations from unlabeled data, they extract limited information relying only on one consistency.

\subsection{Multi-Task Self-Supervised Learning}
As various self-supervised learning approaches have been proposed in diverse domains, multi-task self-supervised learning has been studied to learn general representations. In computer vision, multi-task self-supervised learning methods were proposed for convolutional neural networks (CNNs) and Vision Transformer \cite{dosovitskiy2020image}. \cite{doersch2017multi} combined context prediction \cite{doersch2015unsupervised}, colorization \cite{zhang2016colorful}, exemplar \cite{alexey2015discriminative}, and motion segmentation \cite{pathak2017learning} using CNN architecture. This method explored lasso-regularized combination and harmonizing network inputs for effective multi-task self-supervision. \cite{atito2021sit} trained Vision Transformer to learn visual features by investigating the merits of image reconstruction, rotation prediction, and contrastive learning. A multi-task objective function was derived using an uncertainty weighting approach to mitigate difficulty in optimizing the fixed weights of each task. In natural language processing, \cite{wang2020multi} proposed a multi-task self-supervised learning method for disfluency detection. This method combined tagging to detect the added noisy word and sentence classification to distinguish the original sentences from the ones with grammatical errors. In speech recognition, \cite{ravanelli2020multi} enabled robust speech recognition by jointly learning regression workers to predict speech features and binary workers to capture higher-level information from the speech signal. Finally, in time-series analysis, \cite{saeed2019multi} proposed a multi-task self-supervised learning for human activity detection. It learned time-series representations by integrating multiple binary classifications to recognize various transformations. However, it cannot be generally applied to other time-series data because of its specific application.

\section{Proposed Method}
Our proposed framework aims to learn general representations by training three separate consistencies simultaneously in unlabeled time-series data. We train a model to jointly learn contrastive learning for contextual, temporal, and transformation consistency to achieve our goal. In this section, we describe the proposed framework including the three self-supervised tasks. We followed the procedures in previous works where possible, but some were modified if necessary for a multi-task learning setup.

%Figure 2
\begin{figure*}[!t]
    \centering
    \includegraphics[width=0.7\textwidth]{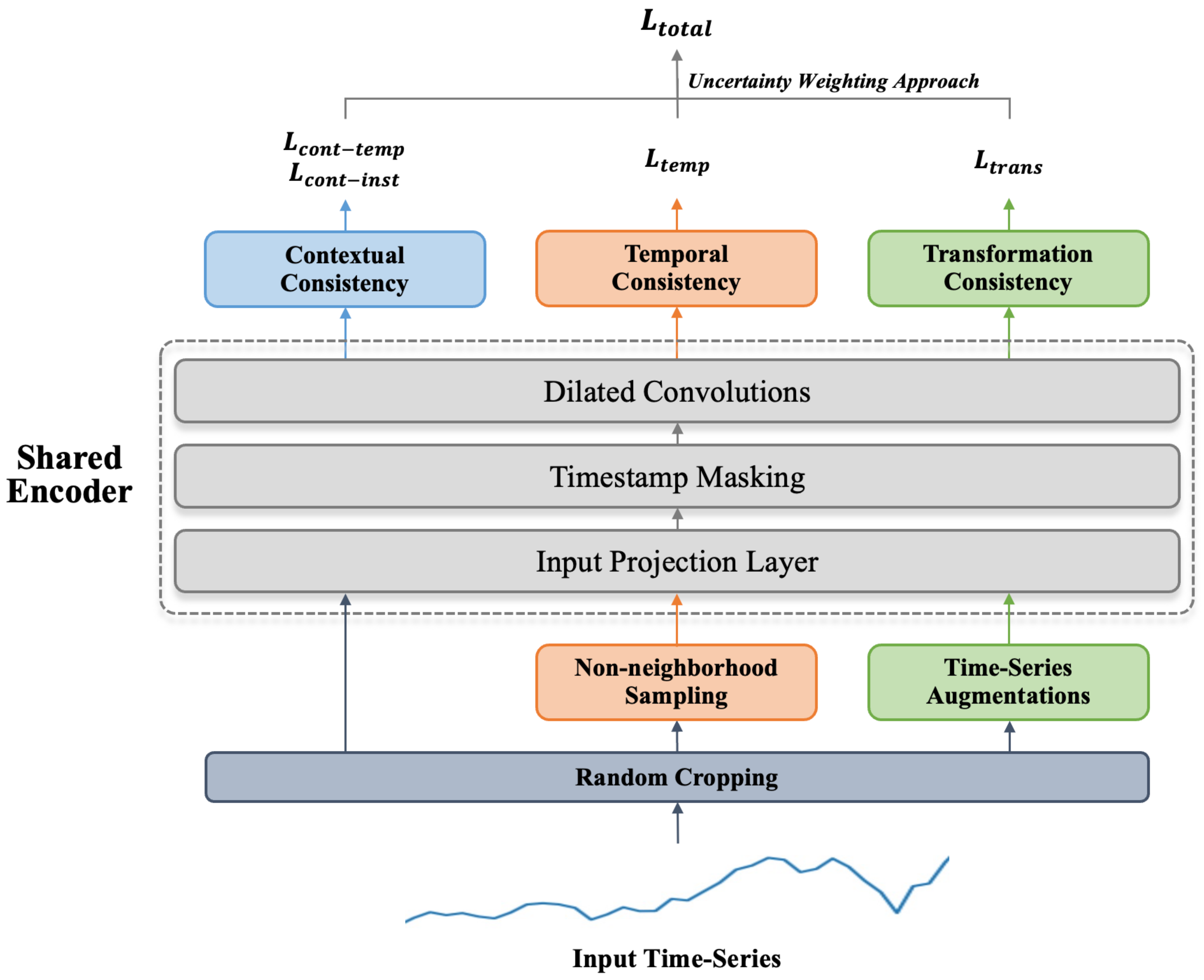}
    \caption{Overall architecture of the multi-task self-supervised time-series representation learning}
    \label{fig2}
\end{figure*}

\subsection{Model Architecture}
Given a set of time series $\mathbf{X}=\{\mathbf{X}_1, \mathbf{X}_2, \dotsm, \mathbf{X}_N\}$ of $N$ instances, we aim to train the shared encoder to map the $i$-th instance $\mathbf{X}_i$ to timestamp-level representation $\mathbf{R}_i$ satisfying three distinct consistencies. Each instance $\mathbf{X}_i=\{\mathbf{x}_{i,1}, \mathbf{x}_{i,2}, \dotsm, \mathbf{x}_{i,T}\}$ constitutes a sequence of $T$ feature vectors $\mathbf{x}_{i,t}\in\mathbb{R}^{m}$, where $m$ is the number of variables. The representation $\mathbf{R}_i=\{\mathbf{r}_{i,1}, \mathbf{r}_{i,2}, \dotsm, \mathbf{r}_{i,T}\}$ has the dimension of $T\times K$ where $K$ is the dimension of representation vector in each time segment.

The overall architecture of multi-task self-supervised time-series representation learning is shown in Figure \ref{fig2}. Given input time-series data, we adopt distinct preprocessing methods to generate positive and negative pairs for each self-supervised task. Specifically, we first sample two overlapping time segments from an input time-series data by random cropping. These segments are commonly used in all tasks, and additional ones are required for temporal and transformation consistency. In temporal consistency, we randomly sample time segments from the non-neighborhood where the time-series data are no longer stationary. In transformation consistency, weak and strong time-series augmentations apply to the overlapping time segments above. Once positive and negative pairs are generated for all tasks, they are fed into the shared encoder. The model then performs each self-supervised task and calculates contrastive loss. Finally, the multi-task loss is derived by effectively combining these losses using an uncertainty weighting approach. This total loss allows the shared encoder to learn general representations by jointly optimizing multiple consistencies together.

In this study, we used the encoder architecture proposed by TS2Vec \cite{yue2022ts2vec}. The shared encoder consists of an input projection layer, a timestamp masking module, and a dilated CNN module. The input projection layer is a fully connected layer that extracts high-dimensional latent features from input time segments. The timestamp masking module masks the latent vectors at randomly selected timestamps from a Bernoulli distribution with $p=0.5$. The dilated CNN module extracts timestamp-level representations by capturing temporal dynamics with a large receptive field. This module consists of ten residual blocks containing two one-dimensional dilated convolutional layers and a GELU activation function with skip connections between two adjacent blocks.

%Figure 3
\begin{figure*}[!t]
    \centering
    \subfloat[Contextual Consistency\label{fig3a}]{%
       \includegraphics[height=7.5cm]{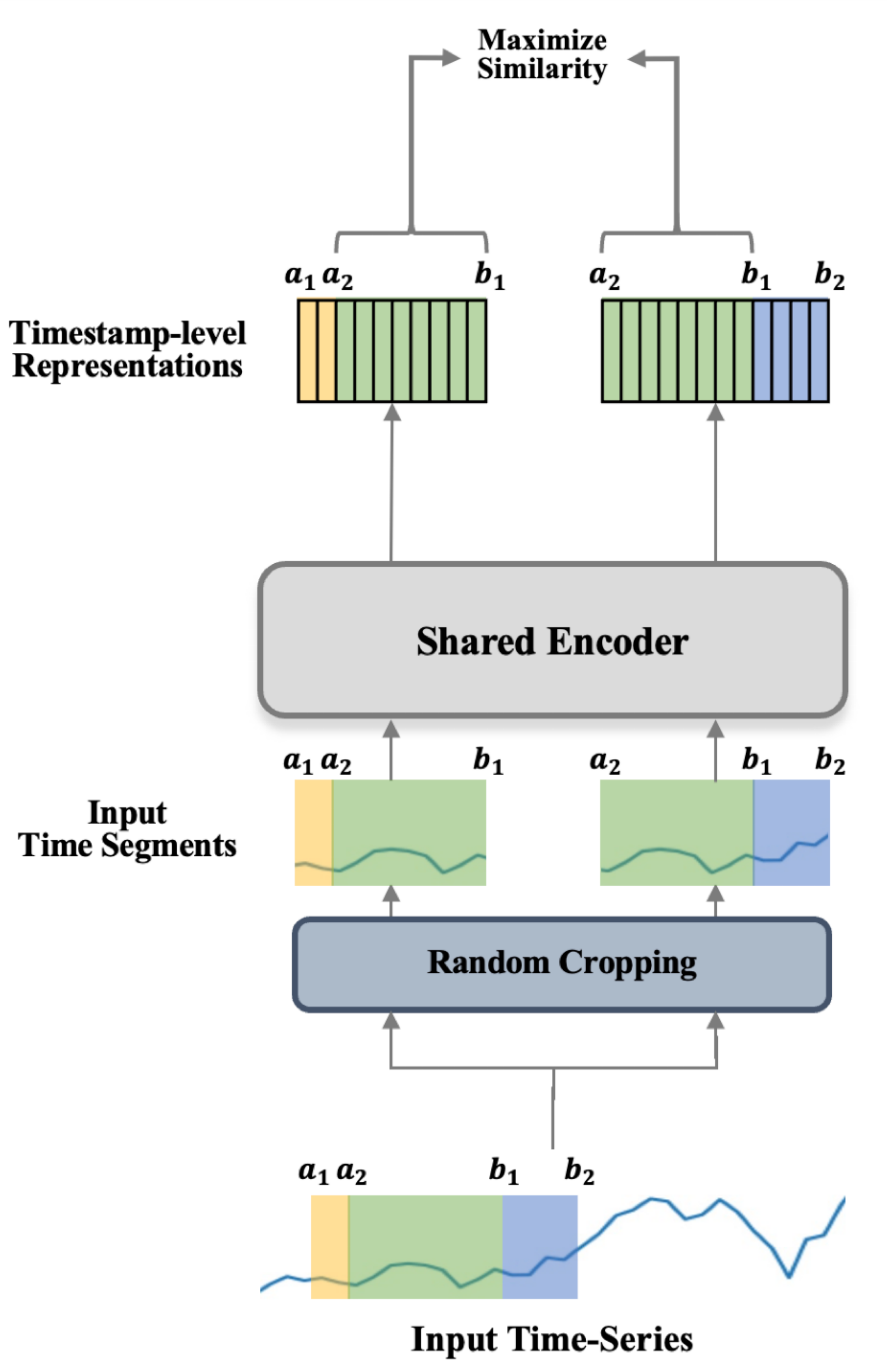}}
    \hfil
    \subfloat[Temporal Consistency\label{fig3b}]{%
        \includegraphics[height=7.5cm]{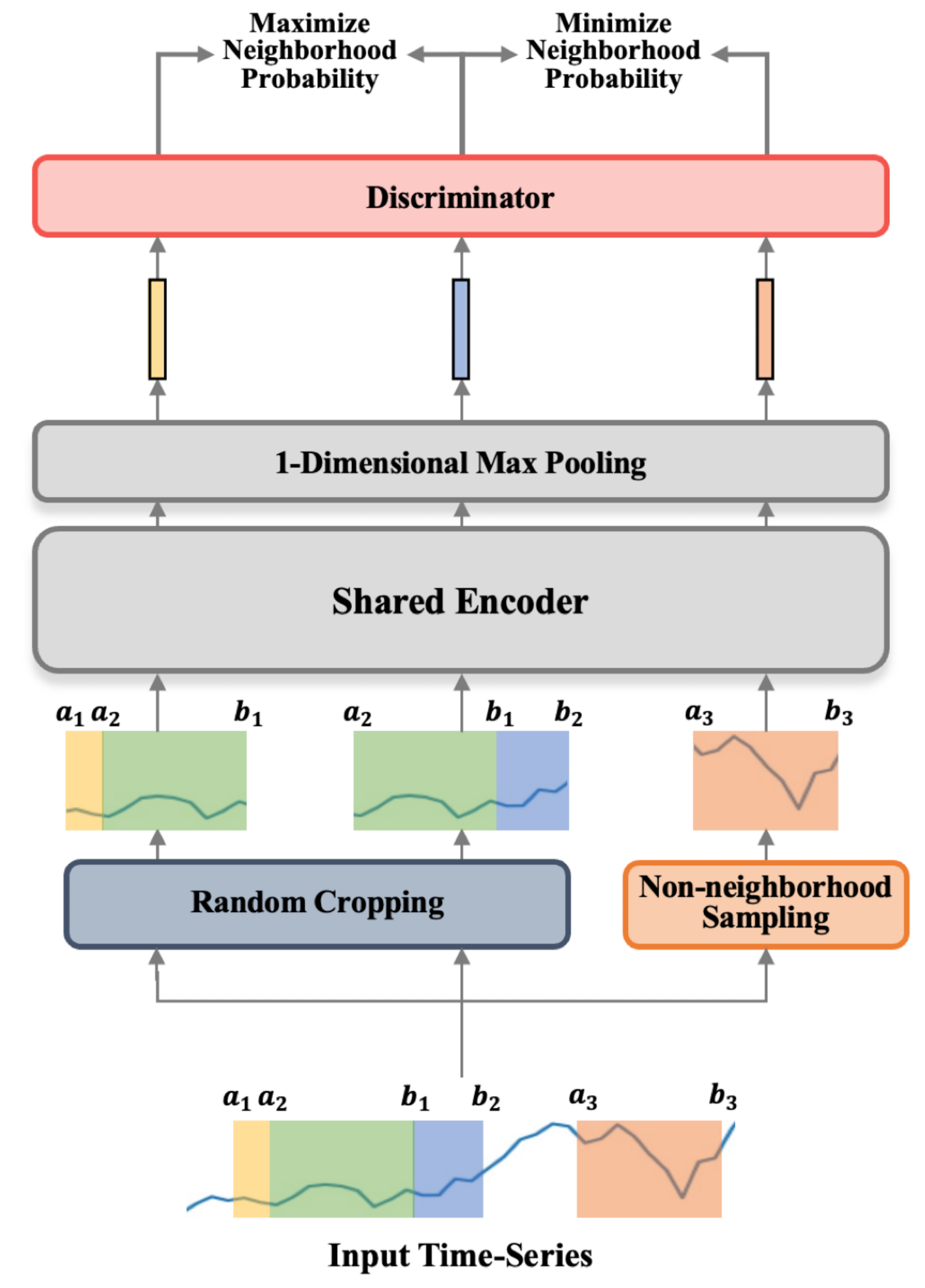}}
    \hfil
    \subfloat[Transformation Consistency\label{fig3c}]{%
        \includegraphics[height=7.5cm]{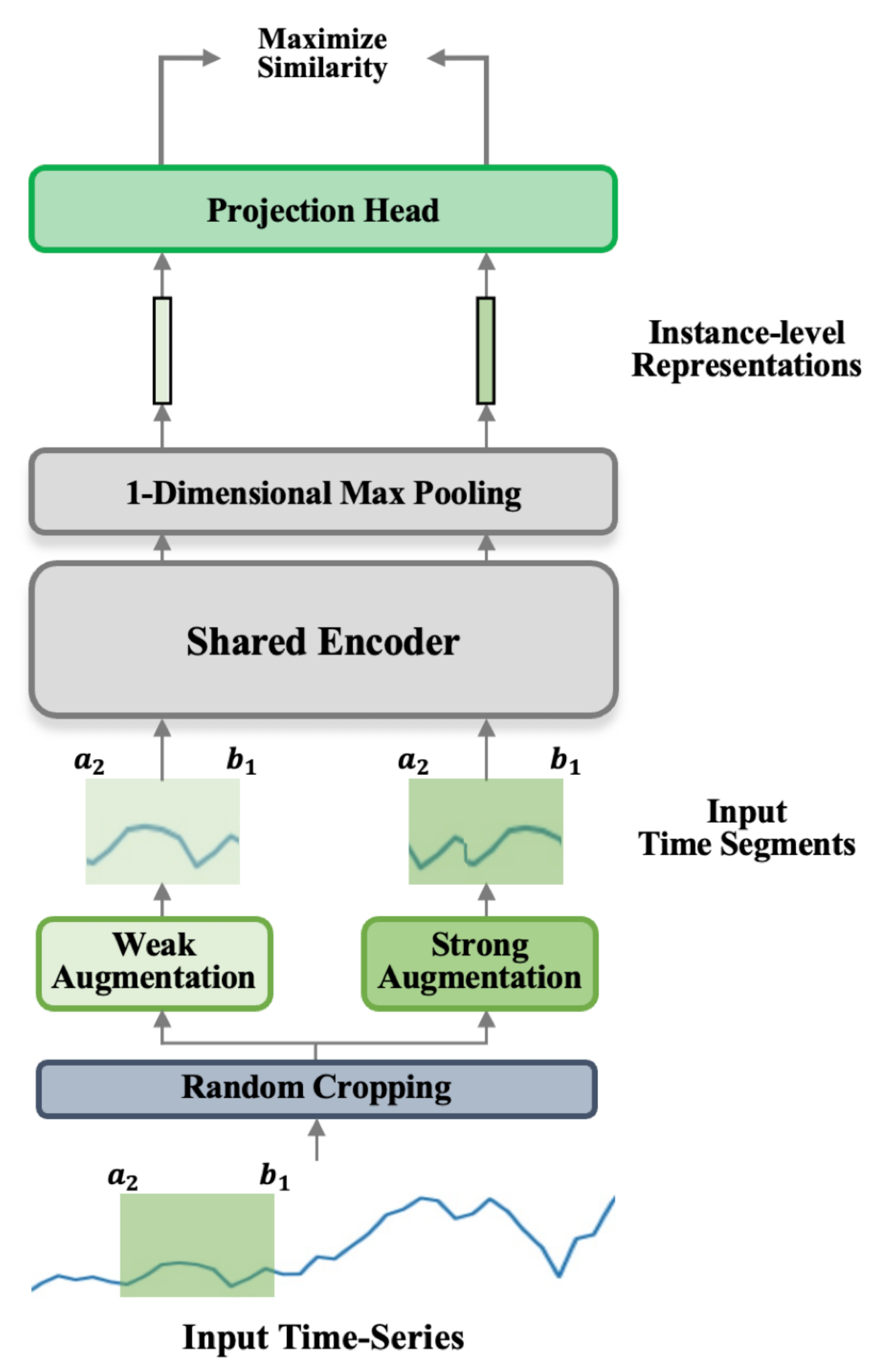}}
    \caption{Contrastive learning for contextual, temporal, and transformation consistency}
    \label{fig3}
\end{figure*}

\subsection{Self-Supervised Tasks}
\subsubsection{Contrastive Learning for Contextual Consistency}\label{sec321}
To encourage contextual consistency in time-series data, we investigate the contrastive learning framework proposed by TS2Vec \cite{yue2022ts2vec}. This self-supervised task performs timestamp-wise and instance-wise contrasting over multiple scales, considering the representations at the same timestamp in two overlapping time segments as positive pairs. The overall structure of contextual consistency is shown in Figure \ref{fig3a}. Given input time-series data $\mathbf{X}_i\in\mathbb{R}^{T\times m}$, random cropping is adapted to generate two overlapping time segments $\mathbf{X}_{i, a_1:b_1}$ and $\mathbf{X}_{i, a_2:b_2}$ under the condition $0<a_1\le a_2\le b_1\le b_2\le T$. For a multi-task setup, we limit the maximum length of the randomly cropped segment to $T\times l$ with hyperparameter $l\in [0, 1]$. In this self-supervised task, these two context views are used as input to ensure that representations of the overlapping timestamps in different contexts are consistent.

Given input segments $\mathbf{X}_{i, a_1:b_1}$ and $\mathbf{X}_{i, a_2:b_2}$, the shared encoder maps them to timestamp-level representations $\mathbf{R}_{i,a_1:b_1}=\{\mathbf{r}_{i,a_1}, \dotsm, \mathbf{r}_{i,b_1}\}$ and $\mathbf{R}^\prime_{i,a_2:b_2}=\{\mathbf{r}^\prime_{i,a_2}, \dotsm, \mathbf{r}^\prime_{i,b_2}\}$, where $[a_2, b_1]$ is an overlapping time region. In the shared encoder, the timestamp masking module generates augmented context views and makes representations at the same timestamp have differing context information. Based on these timestamp-level representations, timestamp-wise and instance-wise contrasting are performed to enforce contextual consistency. Both contrasting methods treat the representations at the same timestamp from two augmented views as positives. Specifically, $\mathbf{r}_{i,t}$ and $\mathbf{r}^\prime_{i,t}$, which are representations at timestamp $t$ from $\mathbf{X}_{i, a_1:b_1}$ and $\mathbf{X}_{i, a_2:b_2}$, are a positive pair. In contrast, different negative samples are chosen depending on the goal of the two contrasting methods. First, timestamp-wise contrasting seeks to make the shared encoder learn discriminative representations over time. Representations at different timestamps from the same context view are considered negative samples to achieve this goal. From these positive and negative samples, the timestamp-wise contrastive loss is defined as follows: 

%Equ 1
\begin{equation}
    L_{cont-temp} = \sum_{i}\sum_{t}{-\log{\frac{\exp\left(\mathbf{r}_{i,t}\cdot \mathbf{r}_{i,t}^\prime\right)}{\sum_{t^\prime\in\left[a_2,b_1\right]}\left(\exp\left(\mathbf{r}_{i,t}\cdot \mathbf{r}_{i,t^\prime}^\prime\right)+\mathbbm{1}_{\left[t\neq t^\prime\right]}\exp\left(\mathbf{r}_{i,t}\cdot \mathbf{r}_{i,t^\prime\ }\right)\right)}}},
    \label{equ:equ1}
\end{equation}
where $\mathbbm{1}$ is the indicator function. Next, instance-wise contrasting aims to learn discriminative representations at the same timestamps over an instance. In this contrasting method, negative samples are representations at the same timestamp of other instances in the same batch whose batch size is $B$. Based on these samples, the instance-wise contrastive loss is formulated as Equation (\ref{equ:equ2}).

%Equ 2
\begin{equation}
    L_{cont-inst}=\sum_{i}\sum_{t}{-\log{\frac{\exp\left(\mathbf{r}_{i,t}\cdot \mathbf{r}_{i,t}^\prime\right)}{\sum_{j=1}^{B}\left(\exp\left(\mathbf{r}_{i,t}\cdot \mathbf{r}_{j,t}^\prime\right)+\mathbbm{1}_{\left[i\neq j\right]}\exp\left(\mathbf{r}_{i,t}\cdot \mathbf{r}_{j,t}\right)\right)}}}
    \label{equ:equ2}
\end{equation}

Finally, these contrastive losses are calculated in a hierarchical framework to force the shared encoder to model contextual consistency at multiple scales. The hierarchical contrasting applies one-dimensional max pooling on the timestamp-level representations along the time axis and computes timestamp-wise and instance-wise losses in Equations (\ref{equ:equ1}) and (\ref{equ:equ2}). Finally, the hierarchical timestamp-wise and instance-wise losses are derived by summing contrastive losses over multiple scales.

\subsubsection{Contrastive Learning for Temporal Consistency}
To leverage temporal consistency in time-series data, we explore the contrastive learning framework proposed by TNC \cite{tonekaboni2021unsupervised}. This self-supervised task attempts to take advantage of the local smoothness by performing temporal neighborhood coding to learn latent space where the neighborhood distribution is distinguishable from the non-neighborhood one. Figure \ref{fig3b} shows the overall structure of temporal consistency. This self-supervised task assumes that time segments in neighborhoods have similar properties while ones in the non-neighborhood are different. To jointly learn temporal and contextual consistency, we consider the two overlapping time segments $\mathbf{X}_{i, a_1:b_1}$ and $\mathbf{X}_{i, a_2:b_2}$, which are generated by random cropping in Section \ref{sec321}, as a neighborhood pair. We then obtain non-neighboring segments of the overlapping region $\mathbf{X}_{i, a_2:b_1}$ using statistical testing proposed by TNC \cite{tonekaboni2021unsupervised}. Specifically, we find the temporal neighborhood of the overlapping region automatically and sample non-neighboring segments outside of this neighborhood. Relying on local smoothness, we assume that the center timestamp $t$ of neighboring segment follows Gaussian distribution $t\sim \mathcal{N}\left((a_2+b_1)/2,\,\eta(b_1-a_2)\right)$. Its mean is the center timestamp of the overlapping region $\mathbf{X}_{i, a_2:b_1}$, and its variance depends on the length of this region $b_1-a_2$ and $\eta$, which adjusts the range of the neighborhood. The temporal neighborhood where stationarity is satisfied is determined by measuring the $p$-value from the Augmented Dickey-Fuller statistical test on the neighborhood area while gradually increasing from $\eta = 1$. Finally, we sample non-neighboring segment $\mathbf{X}_{i, a_3:b_3}$ whose size is $b_1-a_2$, and consider this segment as a negative one because it is likely to be different from overlapping segment $\mathbf{X}_{i, a_2:b_1}$. Once we obtain three input time segments, we constitute a non-neighborhood pair with $\mathbf{X}_{i, a_3:b_3}$ and one segment whose center timestamp is far away from the negative sample among $\mathbf{X}_{i, a_1:b_1}$ and $\mathbf{X}_{i, a_2:b_2}$. For example, in Figure \ref{fig3b}, $\mathbf{X}_{i, a_1:b_1}$ and $\mathbf{X}_{i, a_3:b_3}$ are a non-neighboring pair while $\mathbf{X}_{i, a_1:b_1}$ and $\mathbf{X}_{i, a_2:b_2}$ are a neighboring one. We denote $\mathbf{X}_{i, a_1:b_1}$, $\mathbf{X}_{i, a_2:b_2}$, and $\mathbf{X}_{i, a_3:b_3}$ as $\mathbf{X}^{anc}_{i, a_1:b_1}$, $\mathbf{X}^{N}_{i, a_2:b_2}$, and $\mathbf{X}^{\bar{N}}_{i, a_3:b_3}$ relatively. 

Given neighboring and non-neighboring pairs, the shared encoder extracts timestamp-level representations $\mathbf{R}^{anc}_{i, a_1:b_1}$, $\mathbf{R}^{N}_{i, a_2:b_2}$, and $\mathbf{R}^{\bar{N}}_{i, a_3:b_3}$. To identify the neighboring pair from the non-neighboring one, we extract instance-level representations $\mathbf{r}^{anc}_{i}$, $\mathbf{r}^{N}_{i}$, and $\mathbf{r}^{\bar{N}}_{i}$ by adopting one-dimensional max pooling to timestamp-level ones along the time axis. They have integrated information on timestamp-level representations. We then introduce a discriminator to estimate the probability of a neighboring pair. For the discriminator, we use a multi-headed binary classifier whose goal is to output one if the input representation pair are neighbors and zero otherwise. Contrastive loss for temporal consistency is defined as follows:

%Equ 3
\begin{equation}
    L_{temp}=-\sum_{i}{\log{D\left(\mathbf{r}_i^{anc},\mathbf{r}_i^N\ \right)} \\
    +\left(1-w_{i}\right)\times\log{\left(1-D\left(\mathbf{r}_i^{anc},r_i^{\bar{N}}\right)\right)}
    +w_{i}\times\log{D\left(\mathbf{r}_i^{anc},\mathbf{r}_i^{\bar{N}}\right)}},
    \label{equ:equ3}
\end{equation}
where $w_{i}$ is a parameter to adjust the samples from the non-neighboring region. This parameter is introduced to mitigate the issue of sampling bias using positive-unlabeled (PU) learning. In PU learning, a binary classifier is trained only on a few labeled positive examples and unlabeled samples. PU learning treats the unlabeled data as negative samples with smaller weights to learn the classifier from positive unlabeled input data. Based on this method, we treat each sample from the neighborhood as a positive example with unit weight, while we consider each sample from the non-neighborhood as a combination of a positive example with weight $w$ and a negative example with complementary weight $1-w$. Consequently, this loss forces the shared encoder to learn discriminative representations for the neighborhood and non-neighborhood without sampling bias in contrastive learning. 

\subsubsection{Contrastive Learning for Transformation Consistency}
To explore transformation consistency, we investigate the contrastive learning framework proposed by TS-TCC \cite{eldele2021time}. The goal of this self-supervised task is to encourage the network to learn transform-invariant representations by maximizing agreement between instance-level representations of differently augmented views from a single input. The overall structure of transformation consistency is shown in Figure \ref{fig3c}. For multi-task setup, we use the overlapping region $\mathbf{X}_{i, a_2:b_1}$ between $\mathbf{X}_{i, a_1:b_1}$ and $\mathbf{X}_{i, a_2:b_2}$, which are generated by random cropping in Section \ref{sec321}, as raw input time-series data. We adopt weak and strong augmentations to this raw input and obtain two different augmented segments $\mathbf{X}^W_{i, a_2:b_1}$ and $\mathbf{X}^S_{i, a_2:b_1}$. Specifically, weak augmentation adds random variations to the input and amplifies its magnitude. Strong augmentation divides the input into a random number of sub-series (a maximum of five), shuffles them, and adds a random jittering to the permuted time-series data. The transformed versions $\mathbf{X}^W_{i, a_2:b_1}$ and $\mathbf{X}^S_{i, a_2:b_1}$ are different yet correlated and thus are considered as positive pairs.

Given two augmented views, the shared encoder maps input time segments $\mathbf{X}^W_{i, a_2:b_1}$ and $\mathbf{X}^S_{i, a_2:b_1}$ to timestamp-level representations $\mathbf{R}^W_{i, a_2:b_1}$ and $\mathbf{R}^S_{i, a_2:b_1}$. We obtain instance-level ones $\mathbf{r}^W_{i}$ and $\mathbf{r}^S_{i}$ by applying one-dimensional max pooling to two timestamp-level representations along the time axis. To perform contrastive learning, we introduce the projection head to map $\mathbf{r}^W_{i}$ and $\mathbf{r}^S_{i}$ to the space where the contrastive loss is applied. We use a multi-layer perceptron with two fully-connected layers as projection head. In this self-supervised task, negative samples are built on the batch. Given a batch of $B$ raw input time-series data, it has $2B$ instance-level representations from the augmented views generated by two different augmentations. In this batch, $\mathbf{r}^W_{i}$ and $\mathbf{r}^S_{i}$ from the same input are considered to be positive pairs. In contrast, we treat the remaining $2\left(B - 1\right)$ representations from other inputs within this batch as the negative samples. Contrastive loss for transformation consistency is formulated as follows:

%Equ 4
\begin{equation}
    L_{trans}=\sum_{i}{-\log{\frac{\exp\left(sim\left(\mathbf{r}_i^W,\mathbf{r}_i^S\right)/\tau\right)}{\sum_{j=1}^{2B}{\mathbbm{1}_{\left[i\neq j\right]}\exp\left(sim\left(\mathbf{r}_i^W,\mathbf{r}_j\ \right)/\tau\right)}}}},
    \label{equ:equ4}
\end{equation}
where $sim$ is the cosine similarity function. This contrastive loss enables the shared encoder to explore transformation consistency by enforcing that representations of the two augmented views from the same input are consistent.

\subsection{Multi-Task Self-Supervised Training}
The multi-task self-supervised time-series representation learning is optimized jointly with the contrastive loss for contextual, temporal, and transformation consistency. The objectives of this multi-task contrastive learning are as follows: 1) encouraging consistency of the common sub-series in two augmented contexts, 2) exploring consistency of the adjacent time segments with stationary properties, and 3) modeling consistency between differently augmented views from a single input.

In general, the multi-task loss is defined as a weighted sum of the losses for multiple tasks. However, this requires an expensive process for optimizing the fixed weights of each task. To alleviate this issue, we investigate the uncertainty weighting approach proposed by \cite{kendall2018multi}. This method weighs multiple contrastive losses by considering the homoscedastic uncertainty of each task. It allows the network to be optimized simultaneously by various quantities having different units or scales. The final multi-task loss is formulated as follows:

%Equ 5
\begin{equation}
    \begin{split}
        L_{total}=&\frac{1}{\alpha_1^2}\times\ L_{cont-temp}+\frac{1}{\alpha_2^2}\times\ L_{cont-inst}+\frac{1}{\alpha_3^2}\times\ L_{temp}+\frac{1}{\alpha_4^2}\times\ L_{trans} \\ &+\log{\left(\alpha_1\right)}+\log{\left(\alpha_2\right)}+\log{\left(\alpha_3\right)}+\log{\left(\alpha_4\right)},
    \end{split}
    \label{equ:equ5}
\end{equation}
where the weights $\alpha_1$, $\alpha_2$, $\alpha_3$, and $\alpha_4$ are learnable parameters and represent the contribution of the corresponding contrastive loss. Consequently, we enable efficient multi-task self-supervised learning and thus encourage the network to learn representations exploring various consistencies in time-series data.

\section{Experiments}
\subsection{Experimental Settings}
\subsubsection{Datasets}
We evaluated our proposed model on three distinct downstream tasks: time-series classification, forecasting, and anomaly detection. We conducted experiments on various datasets for each downstream task. First, in time-series classification, the UEA archive\footnote{\url{http://www.timeseriesclassification.com}} \cite{bagnall2018uea} was adopted for evaluation. The UEA archive consists of 30 multivariate time-series classification datasets. We compared the proposed method with the benchmark models on all datasets. Furthermore, we analyzed our model on Human Activity Recognition (HAR)\footnote{\url{https://archive.ics.uci.edu/ml/datasets/human+activity+recognition+using+smartphones}} \cite{anguita2013public} and fault diagnosis datasets. We conducted a qualitative evaluation of the learned representations on the HAR dataset. This dataset is collected from 30 individual subjects who perform six activities: walking, walking upstairs, downstairs, standing, sitting, and lying down. The fault diagnosis dataset\footnote{\url{https://mb.uni-paderborn.de/en/kat/main-research/datacenter/bearing-datacenter/data-sets-and-download}} was adopted to confirm whether the proposed model learns a general representation that can be used in different domains. This dataset consists of four sub-datasets collected under a variety of working conditions, and each sub-dataset includes sensor data and labels for normal and two failures. Next, we evaluated our method on time-series forecasting using the Electricity Transformer Temperature (ETT)\footnote{\url{https://github.com/zhouhaoyi/ETDataset}} \cite{zhou2021informer}. The ETT consists of three datasets, each with six power load features and the target value of oil temperature. For two years, ETTh1 , ETTh2, and ETTm1 are collected every hour and 15 minutes, respectively. Finally, Yahoo\footnote{\url{https://webscope.sandbox.yahoo.com/catalog.php?datatype=s&did=70&guccounter=1&guce_referrer=aHR0cHM6Ly9naXRodWIuY29tL3l1ZXpoaWhhbi90czJ2ZWM& guce_referrer_sig=AQAAAEfu8dEBHF1mI0Bxdt59dIwU4ZaqoLS8HZ1bM7JQVDtAmLmCZuZYwCFCiP8L5lVnD9FIshPchkgnzqn7DojpkRWQ1CM8qY6nrsQ8hNAzCwV3osv-rGIfIqPxx5x1-w6fQbZ28ggajlXfJwp2ahOPxTbkiYmnS37DP8Ygb_24FE97}} \cite{laptevbenchmark} and KPI\footnote{\technical} \cite{ren2019time} datasets were adopted for time-series anomaly detection. The Yahoo dataset contains 367 real and synthetic time-series data related to the Yahoo membership login system. Each data has various outliers, such as change points and outliers. The KPI dataset is a set of multiple minutely sampled KPI curves from several Internet companies.

\subsubsection{Implementation Details}
In this study, we set hyperparameters based on the previous works of the investigated self-supervised tasks. First, the hyperparameters of the shared encoder were set following the TS2Vec \cite{yue2022ts2vec}. The hidden dimension of the input projection layer was set to 64. Each residual block in the dilated CNN module has a kernel size of three and a channel size of 64. The dimension of the time-series representation was set to 320.

The hyperparameters of preprocessing were set for each self-supervised task. We searched the hyperparameter $l$ in $\{0.25, 0.5\}$ in contextual consistency. In temporal consistency, the weight of PU learning was set to 0.05 according to TNC \cite{tonekaboni2021unsupervised}. In transformation consistency, hyperparameters of augmentation were set following TS-TCC \cite{eldele2021time}. Specifically, the scaling ratio for weak augmentation was set to 0.001. In strong augmentation, the jittering ratio was set to 0.001.

Finally, for the training hyperparameters, the batch size and the learning rate were set to eight and 0.001, respectively. The training iteration was set to 200 for datasets with sizes less than 100,000 and 600 otherwise. All experiments were conducted on a Linux workstation with Intel Core i7-9700X CPU, 128 GB RAM, and NVidia GeForce RTX 3090 GPU using PyTorch.

%Table 1
\begin{table*}[t!]
    \caption{Experimental result on multivariate time-series classification}
    \label{table1}
    \centering
    \renewcommand{\arraystretch}{1.3}
    \begin{tabular}{l|ccccccc}
        \hline
        Dataset                   & DTW   & T-Loss               & TNC                  & TS-TCC               & TST                  & TS2Vec               & Ours                 \\ \hline
        ArticularyWordRecognition & 0.987 & 0.943                & 0.973                & 0.953                & 0.977                & {\ul \textbf{0.987}} & {\ul \textbf{0.987}} \\
        AtrialFibrillation        & 0.200 & 0.133                & 0.133                & 0.267                & 0.067                & 0.200                & {\ul \textbf{0.400}} \\
        BasicMotions              & 0.975 & {\ul \textbf{1.000}} & 0.975                & {\ul \textbf{1.000}} & 0.975                & 0.975                & 0.975                \\
        CharacterTrajectories     & 0.989 & 0.993                & 0.967                & 0.985                & 0.975                & {\ul \textbf{0.995}} & 0.993                \\
        Cricket                   & 1.000 & 0.972                & 0.958                & 0.917                & {\ul \textbf{1.000}} & 0.972                & {\ul \textbf{1.000}} \\
        DuckDuckGeese             & 0.600 & 0.650                & 0.460                & 0.380                & 0.620                & {\ul \textbf{0.680}} & 0.520                \\
        EigenWorms                & 0.618 & 0.840                & 0.840                & 0.779                & 0.748                & 0.847                & {\ul \textbf{0.878}} \\
        Epilepsy                  & 0.964 & {\ul \textbf{0.971}} & 0.957                & 0.957                & 0.949                & 0.964                & {\ul \textbf{0.971}} \\
        ERing                     & 0.133 & 0.133                & 0.852                & {\ul \textbf{0.904}} & 0.874                & 0.874                & 0.848                \\
        EthanolConcentration      & 0.323 & 0.205                & 0.297                & 0.285                & 0.262                & {\ul \textbf{0.308}} & 0.304                \\
        FaceDetection             & 0.529 & 0.513                & 0.536                & {\ul \textbf{0.544}} & 0.534                & 0.501                & 0.513                \\
        FingerMovements           & 0.530 & {\ul \textbf{0.580}} & 0.470                & 0.460                & 0.560                & 0.480                & 0.530                \\
        HandMovementDirection     & 0.231 & 0.351                & 0.324                & 0.243                & 0.243                & 0.338                & {\ul \textbf{0.351}} \\
        Handwriting               & 0.286 & 0.451                & 0.249                & 0.498                & 0.225                & 0.515                & {\ul \textbf{0.515}} \\
        Heartbeat                 & 0.717 & 0.741                & 0.746                & {\ul \textbf{0.751}} & 0.746                & 0.683                & 0.722                \\
        JapaneseVowels            & 0.949 & {\ul \textbf{0.989}} & 0.978                & 0.930                & 0.978                & 0.984                & 0.984                \\
        Libras                    & 0.870 & {\ul \textbf{0.883}} & 0.817                & 0.822                & 0.656                & 0.867                & {\ul \textbf{0.883}} \\
        LSST                      & 0.551 & 0.509                & {\ul \textbf{0.595}} & 0.474                & 0.408                & 0.537                & 0.553                \\
        MotorImagery              & 0.500 & 0.580                & 0.500                & {\ul \textbf{0.610}} & 0.500                & 0.510                & 0.500                \\
        NATOPS                    & 0.883 & 0.917                & 0.911                & 0.822                & 0.850                & {\ul \textbf{0.928}} & {\ul \textbf{0.928}} \\
        PEMS-SF                   & 0.711 & 0.676                & 0.699                & 0.734                & 0.740                & 0.682                & {\ul \textbf{0.757}} \\
        PenDigits                 & 0.977 & 0.981                & 0.979                & 0.974                & 0.560                & 0.989                & {\ul \textbf{0.990}} \\
        PhonemeSpectra            & 0.151 & 0.222                & 0.207                & {\ul \textbf{0.252}} & 0.085                & 0.233                & 0.248                \\
        RacketSports              & 0.803 & {\ul \textbf{0.855}} & 0.776                & 0.816                & 0.809                & {\ul \textbf{0.855}} & 0.836                \\
        SelfRegulationSCP1        & 0.775 & {\ul \textbf{0.843}} & 0.799                & 0.823                & 0.754                & 0.812                & {\ul \textbf{0.843}} \\
        SelfRegulationSCP2        & 0.539 & 0.539                & 0.550                & 0.533                & 0.550                & {\ul \textbf{0.578}} & 0.567                \\
        SpokenArabicDigits        & 0.963 & 0.905                & 0.934                & 0.970                & 0.923                & {\ul \textbf{0.988}} & 0.981                \\
        StandWalkJump             & 0.200 & 0.333                & 0.400                & 0.333                & 0.267                & 0.467                & {\ul \textbf{0.533}} \\ 
        UWaveGestureLibrary       & 0.903 & 0.875                & 0.759                & 0.753                & 0.575                & 0.906                & {\ul \textbf{0.916}} \\
        InsectWingbeat            & -     & 0.156                & {\ul \textbf{0.469}} & 0.264                & 0.105                & 0.466                & 0.458                \\ \hline
        On the first 29 datasets:& \multicolumn{7}{c}{} \\
        \quad Total Best Acc       & 0     & 7                    & 1                    & 6                    & 1                    & 8                    & {\ul \textbf{14}}    \\
        \quad Average Acc          & 0.650 & 0.675                & 0.677                & 0.682                & 0.635                & 0.712                & {\ul \textbf{0.725}} \\
        \quad Average Rank         & 4.310 & 3.586                & 4.448                & 4.276                & 4.897                & 2.828                & {\ul \textbf{2.276}} \\ \hline
    \end{tabular}
\end{table*}

\subsection{Experimental Results}
\subsubsection{Time-Series Classification} \label{sec421}
In this section, we evaluated our proposed method in time-series classification using the UEA archive. We followed the experimental protocol of TS2Vec \cite{yue2022ts2vec}. The representations were extracted from the pretrained representation models on the training data. With regard to the timestamp-level representation, instance-level representation was derived by applying one-dimensional max pooling because the datasets in the UEA archive have the classes labeled on the instance. We trained RBF kernel-based support vector machine on the instance-level representations of training data and obtained the predicted classes for test data. The penalty for the support vector machine was selected using grid search by cross-validation in the search space $\{10^{-4},10^{-3},\cdots,10^3,10^4,\infty\}$. In this study, we compared our method with the benchmark models, including DTW \cite{chen2013dtw}, T-Loss \cite{franceschi2019unsupervised}, TNC\cite{tonekaboni2021unsupervised}, TS-TCC\cite{eldele2021time}, TST \cite{zerveas2021transformer}, and TS2Vec \cite{yue2022ts2vec} for 30 classification datasets in the UEA archive.

%Figure 4
\begin{figure*}[!t]
    \centering
    \subfloat[BasicMotions - Dataset\label{fig4a}]{%
       \includegraphics[height=5cm]{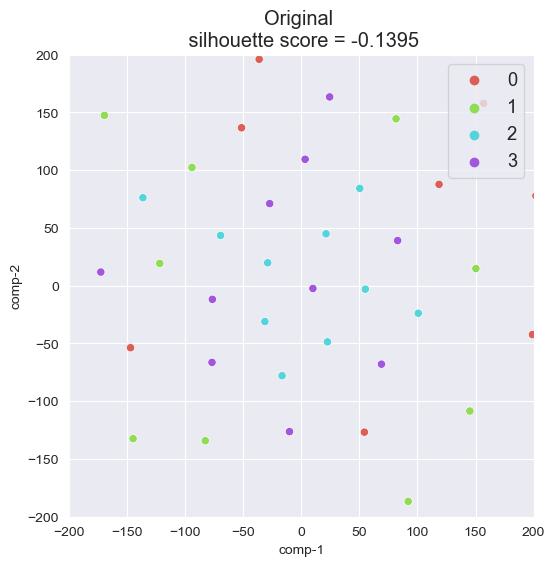}}
    \subfloat[BasicMotions - TS2Vec\label{fig4b}]{%
        \includegraphics[height=5cm]{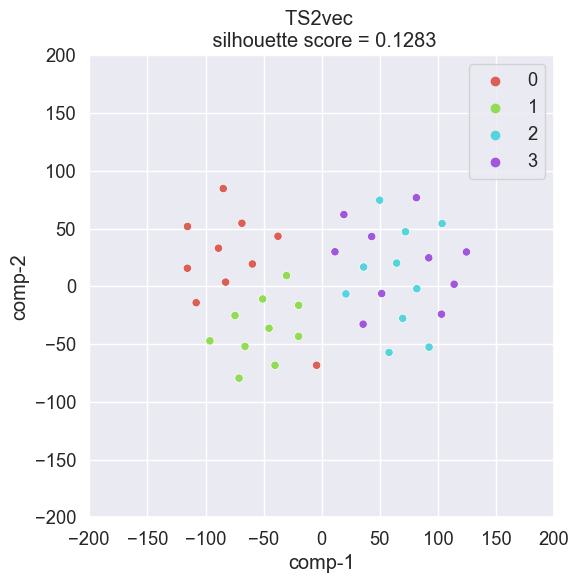}}
    \subfloat[BasicMotions - Ours\label{fig4c}]{%
        \includegraphics[height=5cm]{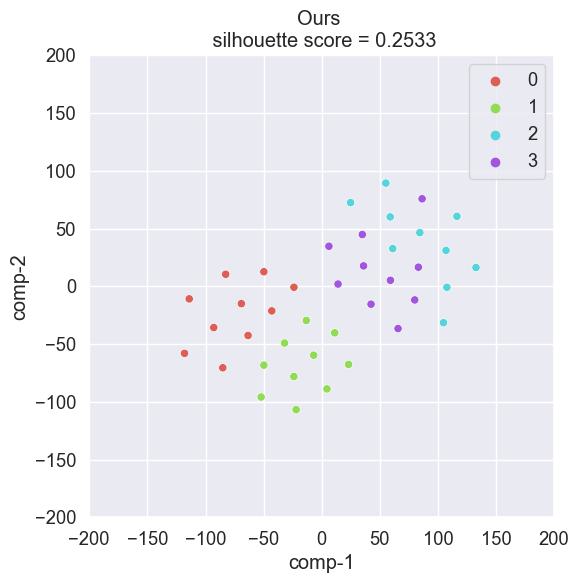}}
    \vfil
    \subfloat[RacketSports - Dataset\label{fig4d}]{%
       \includegraphics[height=5cm]{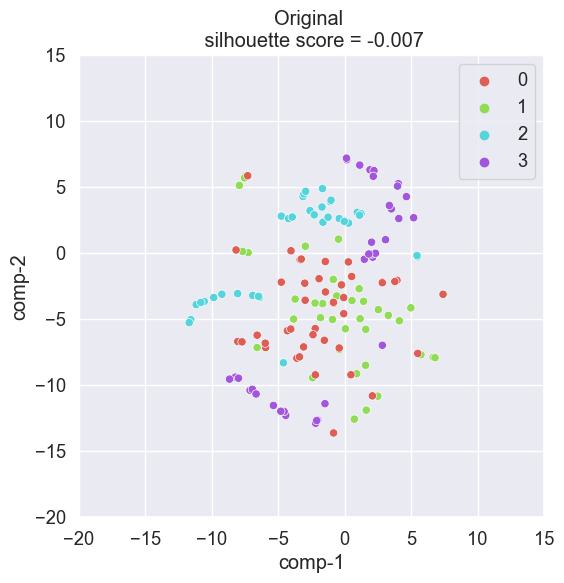}}
    \subfloat[RacketSports - TS2Vec\label{fig4e}]{%
        \includegraphics[height=5cm]{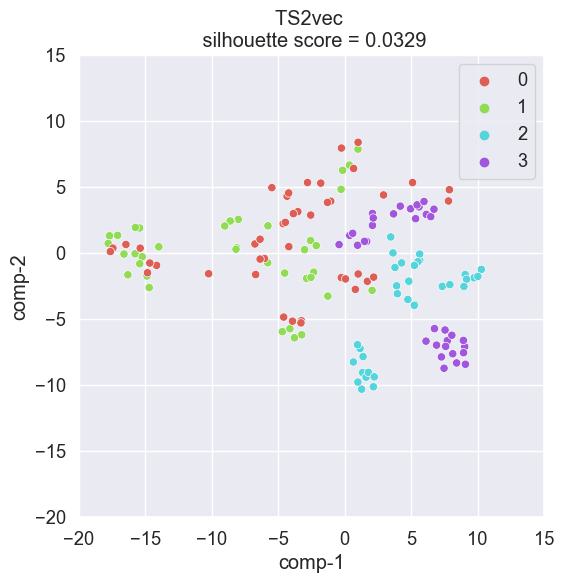}}
    \subfloat[RacketSports - Ours\label{fig4f}]{%
        \includegraphics[height=5cm]{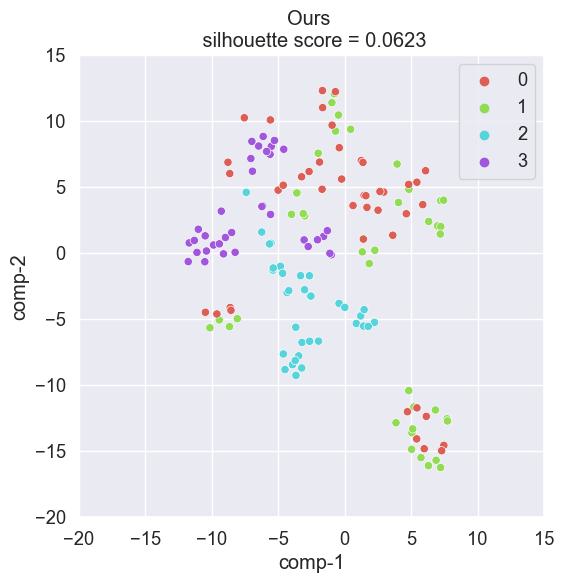}}
    \caption{Visualization of learned representations in BasicMotions and RacketSports datasets}
    \label{fig4}
\end{figure*}

The overall experimental results are presented in Table \ref{table1}, and the best performances are highlighted in bold and underlined. We reported the classification accuracy of benchmark models from their original papers and TS2Vec \cite{yue2022ts2vec}. In the case of the DTW, performance was not recorded in the InsectWingbeat dataset, so average accuracy and rank were calculated based on the remaining 29 datasets. The proposed model outperformed the benchmark models in average accuracy. Specifically, the performance of our model was far higher than that of benchmark models on AtrialFibrillation and StandWalkJump datasets. The proposed model achieved the average rank of 2.276 and the highest accuracy in 14 among 29 datasets. Among the benchmark models, the average accuracy was high in the order of TS2Vec (0.712), TS-TCC (0.682), and TNC (0.677). An interesting observation is that these methods are the representative model for temporal, transformation, and contextual consistency, respectively. Based on this result, we can conclude that jointly learning various self-supervised tasks can be more effective for extracting meaningful knowledge from raw time-series data compared to models using a single consistency.

Next, we visualized the learned representations for qualitative evaluation using t-SNE \cite{van2008visualizing}. We also conducted a quantitative evaluation using the Silhouette score for the classes in the t-SNE space. This score can measure how similar each representation is to its class compared to other classes. Figure \ref{fig4} shows the plots and Silhouette scores of the original dataset and the representations learned from TS2Vec and ours for BasicMotions and RacketSports datasets on the UEA archive. In the BasicMotions dataset, representations learned from our model were located close to data with the same class and far from ones with other classes. Our Silhouette score was significantly higher than that of the original dataset and TS2Vec. These results indicated that the proposed method could learn good representations both qualitatively and quantitatively. Specifically, the original dataset derived a negative Silhouette score, meaning the raw data in the same class were scattered. This was also confirmed in Figure \ref{fig4a}. As shown in Figure \ref{fig4b}, TS2Vec extracted features that can distinguish the overall classes, but these features lacked information to distinguish the second and third classes. TS2Vec also derived a Silhouette score less than half of ours. In the RacketSports dataset, both TS2Vec and our model learned the discriminative representations rather than the original dataset in which data of all classes were aggregated. However, our model yielded a Silhouette score twice as large as that of TS2Vec. From these quantitative and qualitative results, we can confirm that the proposed multi-task representation learning method extracts useful information for classification from data in which information between classes is not distinguished.

\subsubsection{Time-Series Forecasting}
We conducted experiments on three time-series forecasting datasets ETTh1, ETTh2, and ETTm1 following the experimental protocol of TS2Vec \cite{yue2022ts2vec}. In these experiments, multivariate time-series forecasting predicted future data with a length of $H$ based on the representation at the last timestamp of input time-series data. We used ridge regression as a forecasting model with regularization term $\alpha$ selected by a grid search for $\{0.1, 0.2, 0.5, 1, 2, 5, 10, 20, 50, 100, 200, \linebreak[1] 500, 1000\}$. The mean squared error (MSE) and mean absolute error (MAE) were used as evaluation metrics. In this study, we compared our method with the benchmark models, including LSTnet \cite{lai2018modeling}, TCN \cite{bai2018empirical}, LogTrans \cite{li2019enhancing}, StemGNN \cite{cao2020spectral}, Informer \cite{zhou2021informer}, and TS2Vec \cite{yue2022ts2vec}. We confirmed the performance on various $H$. For $H$ larger than 672, the performance of StemGNN was not recorded because it faced an out-of-memory problem even when the batch size was one. Therefore, average MAE and rank were derived except for StemGNN.

%Table 2
\begin{table*}[t!]
    \caption{Experimental result on multivariate time-series forecasting}
    \label{table2}
    \centering
    \renewcommand{\arraystretch}{1.3}
        \begin{tabular}{c|c|ccccccc}
        \hline
        \multirow{2}{*}{Dataset} & \multirow{2}{*}{H} & \multicolumn{7}{c}{MAE}                                                                                                              \\ \cline{3-9}
                         &                    & LSTNet & TCN                  & LogTrans & StemGNN              & Informer             & TS2Vec               & Ours                 \\ \hline
        \multirow{5}{*}{ETTh1}   & 24                 & 0.901  & 0.612                & 0.604    & 0.571                & 0.549                & 0.534                & {\ul \textbf{0.517}} \\
                         & 48                 & 0.96   & 0.617                & 0.757    & 0.618                & 0.625                & 0.555                & {\ul \textbf{0.540}} \\
                         & 168                & 1.214  & 0.738                & 0.846    & {\ul \textbf{0.608}} & 0.752                & 0.636                & 0.620                \\
                         & 336                & 1.369  & 0.800                & 0.952    & 0.73                 & 0.873                & 0.717                & {\ul \textbf{0.698}} \\
                         & 720                & 1.38   & 1.311                & 1.291    & -                    & 0.896                & {\ul \textbf{0.790}} & 0.796 \\ \hline
        \multirow{5}{*}{ETTh2}   & 24                 & 1.457  & 0.888                & 0.750    & 0.883                & 0.665                & {\ul \textbf{0.461}} & 0.495 \\
                                 & 48                 & 1.687  & 0.960                & 1.034    & 0.847                & 1.001                & {\ul \textbf{0.573}} & 0.597 \\
                                 & 168                & 2.513  & 1.407                & 1.681    & 1.228                & 1.515                & 1.065                & {\ul \textbf{1.001}} \\
                                 & 336                & 2.591  & 1.481                & 1.763    & 1.351                & 1.340                & 1.215                & {\ul \textbf{1.144}} \\
                                 & 720                & 3.709  & 1.588                & 1.552    & -                    & 1.473                & 1.373                & {\ul \textbf{1.246}} \\ \hline
        \multirow{5}{*}{ETTm1}   & 24                 & 1.17   & 0.374                & 0.412    & 0.57                 & {\ul \textbf{0.369}} & 0.436                & 0.427                \\
                                 & 48                 & 1.215  & {\ul \textbf{0.450}} & 0.583    & 0.628                & 0.503                & 0.515                & 0.509                \\
                                 & 96                 & 1.542  & 0.602                & 0.792    & 0.624                & 0.614                & 0.549                & {\ul \textbf{0.534}} \\
                                 & 288                & 2.076  & 1.351                & 1.320    & 0.683                & 0.786                & 0.609                & {\ul \textbf{0.591}} \\
                                 & 672                & 2.941  & 1.467                & 1.461    & -                    & 0.926                & 0.655                & {\ul \textbf{0.644}} \\ \hline
        \multicolumn{2}{l|}{Total Best MAE}                & 0      & 1                    & 0        & 1                    & 1                    & 3                    & {\ul \textbf{9}}    \\
        \multicolumn{2}{l|}{Average MAE}       & 1.782  & 0.976                & 1.053    & -                    & 0.859                & 0.712                & {\ul \textbf{0.690}} \\
        \multicolumn{2}{l|}{Average Rank}              & 6.800  & 4.200                & 5.067    & -                    & {\ul \textbf{3.600}} & 2.200                & {\ul \textbf{1.600}} \\ \hline
        \end{tabular}
\end{table*}

Table \ref{table2} presents the overall experimental results on multivariate time-series forecasting with the performance of benchmark models from their original papers and TS2Vec. The best performances are highlighted in bold and underlined. The proposed model showed the lowest average MAE and rank on all ETT datasets. Our model also showed robust performance for the prediction length $H$. Furthermore, especially in long-term prediction, the proposed model outperformed the TS2Vec, a time-series representation learning model. From these results, it can be concluded that the proposed model is an effective time-series representation learning model for long-sequence time-series forecasting.

\subsubsection{Time-Series Anomaly Detection}
We evaluated our proposed model in time-series anomaly detection on the experimental protocol of TS2Vec \cite{yue2022ts2vec}. In this experiment, the representation at the last timestamp in input time-series data was used to determine whether an outlier exists in input time-series data. Following the TS2Vec, the anomaly score was defined as the L1 distance between the representations before and after timestamp masking. F1 score, precision, and recall were used for evaluation metrics. We compared the proposed model with the benchmark methods, including SPOT, DSPOT \cite{siffer2017anomaly}, DONUT \cite{xu2018unsupervised}, SR \cite{ren2019time}, and TS2Vec \cite{yue2022ts2vec}.

The overall experimental results are presented in Table \ref{table3}, and the best performances are highlighted in bold and underlined. We reported the performances of benchmark models from their original papers and TS2Vec. The proposed model showed the highest F1 score in Yahoo and KPI datasets. These results showed that our method is effective in time-series anomaly detection as well as classification and forecasting. Therefore, we can conclude that the proposed model learns general representations that can be utilized in various downstream tasks by combining multiple self-supervised tasks.

\subsection{Analysis}
\subsubsection{Effects of Model Components}
To evaluate the effects of the model components, we conducted experiments for the many variants of our method in time-series classification. We also conducted paired t-test on the performances before and after adding the model component. The $p$-value showed whether the performance significantly improved on 30 datasets in the UEA archive when the model component was added.

%Table 3
\begin{table*}[t!]
    \caption{Experimental results on time-series anomaly detection}
    \label{table3}
    \centering
    \renewcommand{\arraystretch}{1.3}
        \begin{tabular}{c|ccc|cccc}
        \hline
        \multirow{2}{*}{} & \multicolumn{3}{c}{Yahoo}                 & \multicolumn{3}{c}{KPI}                   \\ \cline{2-7}
                          & F1 score             & Precision & Recall & F1 score             & Precision & Recall \\ \hline
        SPOT              & 0.338                & 0.269     & 0.454  & 0.217                & 0.786     & 0.126  \\
        DSPOT             & 0.316                & 0.241     & 0.458  & 0.521                & 0.623     & 0.447  \\
        DONUT             & 0.026                & 0.013     & 0.825  & 0.347                & 0.371     & 0.326  \\
        SR                & 0.563                & 0.451     & 0.747  & 0.622                & 0.647     & 0.598  \\
        TS2Vec            & 0.745                & 0.729     & 0.762  & 0.677                & 0.929     & 0.533  \\
        \hline
        Ours              & {\ul \textbf{0.754}} & 0.762     & 0.747  & {\ul \textbf{0.678}} & 0.941     & 0.530 \\ \hline
        \end{tabular}
\end{table*}

Table \ref{table4} presents the results for each variant on the UEA archive. Temp, Cont, Trans, and UW refer to temporal consistency, contextual consistency, transformation consistency, and uncertainty weighting approach, respectively. The average accuracy increased after adding each self-supervised task and uncertainty weighting sequentially. Furthermore, each model component resulted in a significant performance improvement in all datasets with the low $p$-values. From these results, we can conclude that each self-supervised task helps the network learn better time-series representations. It can also be concluded that the uncertainty weighting approach is effective in the stability of multi-task learning.

%Table 4
\begin{table*}[t!]
    \caption{Effectiveness of the model components}
    \label{table4}
    \centering
    \renewcommand{\arraystretch}{1.3}
        \begin{tabular}{l|c|c}
        \hline
        \multicolumn{1}{c|}{Model variation} & Average Acc & $p$-value        \\ \hline
        Temp + Cont + Trans + UW (ours)      & 0.716       & -                \\ 
        Temp + Cont + Trans                  & 0.704       & 0.006            \\ 
        Temp + Cont                          & 0.691       & 0.006            \\ 
        Temp                                 & 0.670       & 0.064            \\ \hline
        \end{tabular}
\end{table*}

%Table 5
\begin{table*}[t!]
    \caption{Experimental results on cross-domains transfer learning experiment}
    \label{table5}
    \centering
    \renewcommand{\arraystretch}{1.3}
        \begin{tabular}{cccccc}
        \hline
        \multicolumn{2}{c|}{\multirow{2}{*}{}}       & \multicolumn{4}{c}{Target}    \\ \cline{3-6}
        \multicolumn{2}{c|}{}                                                 & A     & B     & C     & D     \\ \hline
        \multicolumn{1}{c|}{\multirow{4}{*}{Source}} & \multicolumn{1}{c|}{A} & 0.994 & 1.000 & 0.989 & 0.999 \\
        \multicolumn{1}{c|}{}                        & \multicolumn{1}{c|}{B} & 0.876 & 1.000 & 1.000 & 0.996 \\
        \multicolumn{1}{c|}{}                        & \multicolumn{1}{c|}{C} & 0.805 & 1.000 & 1.000 & 0.999 \\
        \multicolumn{1}{c|}{}                        & \multicolumn{1}{c|}{D} & 0.806 & 1.000 & 0.999 & 0.997 \\ \hline
        \end{tabular}
\end{table*}

\subsubsection{Visualization}
To understand the learned representations, we conducted qualitative and quantitative evaluations based on the HAR dataset. We compared the representations derived from four models: 1) random initialization, 2) representation learning, 3) fine-tuning, and 4) supervised learning. First, the random initialization is the initialized shared encoder before training. Second, representation learning is the model trained by the proposed multi-task self-supervised time-series representation learning. Third, fine-tuning is the fine-tuned model from the pretrained encoder with labeled data for the classification task. Finally, supervised learning is the encoder trained on the labeled data from scratch. For these models, we visualized the learned representations for qualitative evaluation using t-SNE and conducted a quantitative evaluation using the Silhouette score as shown in Section \ref{sec421}.

%Figure 5
\begin{figure*}[!t]
    \centering
    \subfloat[Random Initialization\label{fig5a}]{%
       \includegraphics[height=6cm]{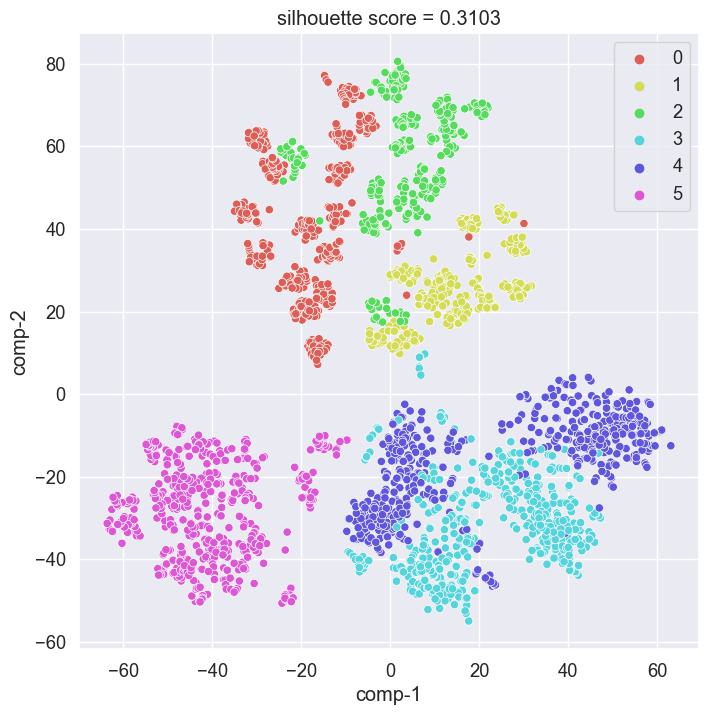}}
    \hfil
    \subfloat[Representation Learning\label{fig5b}]{%
        \includegraphics[height=6cm]{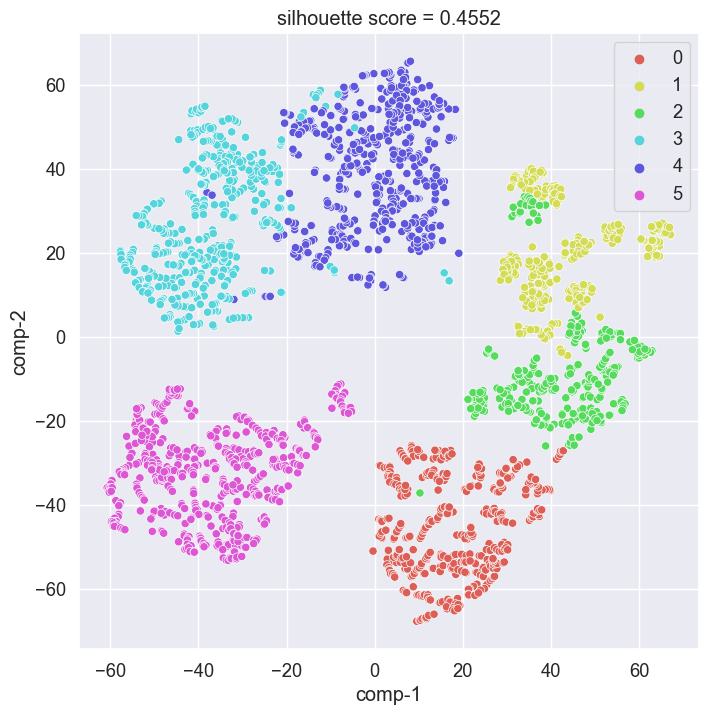}}
    \vfil
    \subfloat[Fine-tuning\label{fig5c}]{%
        \includegraphics[height=6cm]{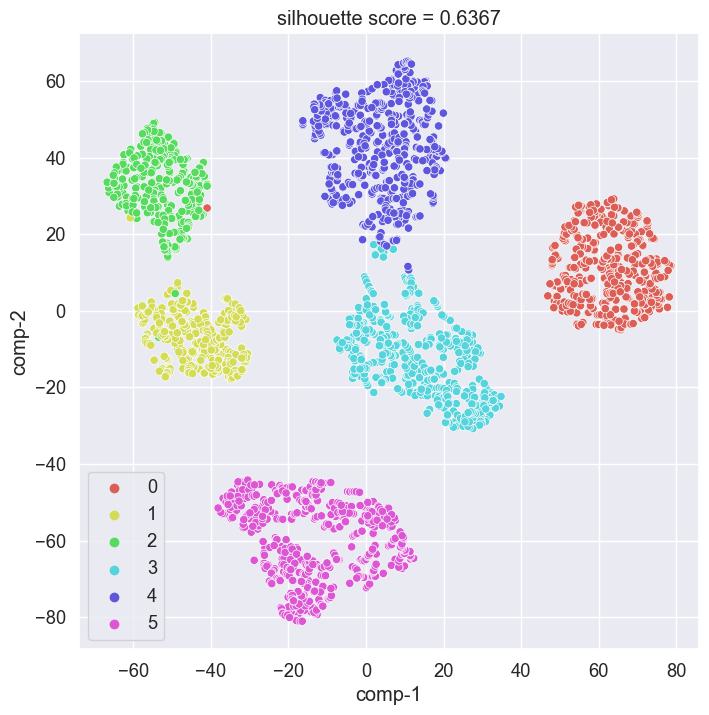}}
    \hfil
    \subfloat[Supervised Learning\label{fig5d}]{%
        \includegraphics[height=6cm]{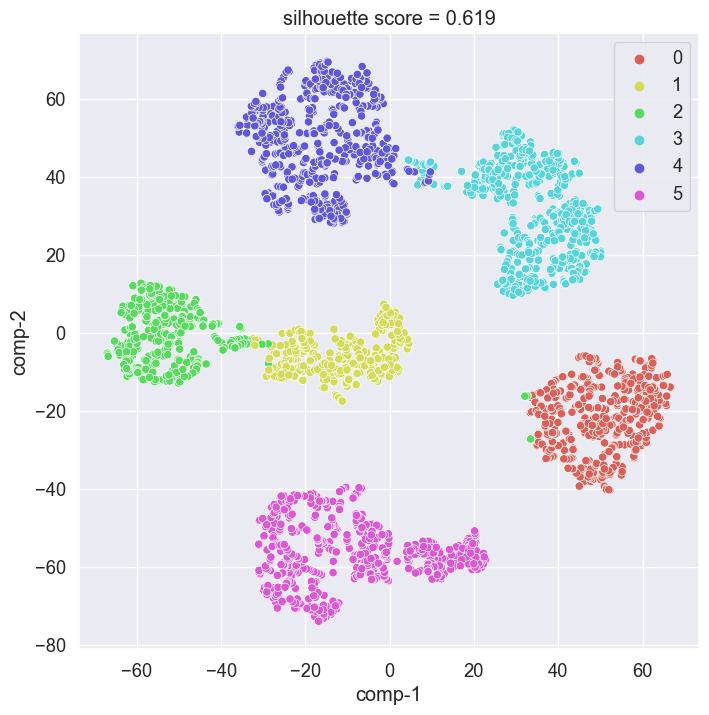}}
    \caption{Visualization of learned representations}
    \label{fig5}
\end{figure*}

Figure \ref{fig5} shows the learned representations of the test dataset derived from the four models. As shown in Figure \ref{fig5a} and \ref{fig5b}, the proposed model can closely map instances with the same class having similar characteristics. Figures \ref{fig5b} and \ref{fig5c} show that the representations became more suitable for the classification task after fine-tuning. The Silhouette score increases in the order of random initialization, representation learning, and fine-tuning. From these results, we can quantitatively confirm that a more suitable representation for the downstream task is learned after representation learning and fine-tuning sequentially. Furthermore, Figure \ref{fig5c} shows slightly more discriminative representations than Figure \ref{fig5d}. Specifically, as shown in Figure \ref{fig5c}, the representation learning learned classification-specific representations rather than supervised learning for classes from zero to two that were not well distinguished in random initialization. These results can also be confirmed quantitatively in the Silhouette score.

\subsubsection{Transfer Learning}
We evaluated the transferability of the proposed method to verify whether it learns general representations that can be adapted to various domains. A transfer learning experiment was conducted on the fault diagnosis dataset collected from four domains. In this experiment, we pretrained the encoder using the unlabeled training data in the source domain and then fine-tuned the pretrained model on the labeled data in the target domain. Accuracy was derived from the test data in the target domain using the fine-tuned model. Table \ref{table5} presents the transferability of the proposed model. Except for the case where the target domain is A, representations trained on source data were adopted to target data satisfactorily. From these results, we can conclude that our method learns general representations useful for diverse domains.

\section{Conclusion}
Time-series analysis helps to extract useful information from large-scale unlabeled time-series data in various fields. However, it is difficult to achieve good performance in the time-series analysis using time-series data with complex dynamics and sparse annotation. To address this issue, self-supervised time-series representation learning has been studied. Most existing methods are based on contrastive learning that maximizes agreement between positive pairs. They attempt to explore different consistency in time-series data using differently defined positive pair and contrastive loss. However, existing methods can limit the knowledge in learned representations because they explore only one consistency for learning time-series representation.

In this paper, we propose a multi-task self-supervised time-series representation learning that combines contrastive learning for contextual, temporal, and transformation consistency. The proposed model can learn general representations by simultaneously encouraging multiple consistencies in time-series data. Contextual consistency treats overlapping timestamps in different context views as positive pairs and allows their representations to be consistent. Temporal consistency leverages the local smoothness of the representations by choosing two adjacent time segments that satisfy stationarity as a positive pair. Transformation consistency learns transformation-invariant representations by enforcing representations of similar samples to be close to each other. Finally, these three self-supervised tasks are jointly optimized using the uncertainty weighting approach for effective multi-task learning. We conducted experiments on downstream tasks to confirm that our method can learn general features for various tasks. The experimental results showed that the proposed model achieved good performance on time-series classification, forecasting, and anomaly detection. We also conducted transfer learning experiments and demonstrated that the proposed model could also learn general representations for various domains.

%Bibliography
\bibliographystyle{unsrt}  
\bibliography{references}

\end{document}